\documentclass[final,1p,times]{elsarticle}

\usepackage{natbib}
\usepackage{hyperref}

\usepackage{times}
\usepackage{amsmath,amssymb,amsfonts}
\usepackage{algorithmic}
\usepackage{subfig}
\usepackage{graphicx}
\usepackage{textcomp}
\usepackage[dvipsnames]{xcolor}
\usepackage{lipsum}
\usepackage{booktabs, multirow}
\usepackage[utf8]{inputenc}
\usepackage[T1]{fontenc}
\usepackage{todonotes}
\usepackage{setspace}

\newcommand{\etal}{\emph{et al.}}
\newcommand{\sota}{state-of-the-art }

\usepackage{filecontents}
\usepackage{tikz}
\usepackage{tikz-qtree}
\usetikzlibrary{trees} 

\usepackage{pdflscape}

\usepackage{newunicodechar}
\newunicodechar{ﬁ}{fi}
\newunicodechar{ﬀ}{ff}

\usetikzlibrary{arrows.meta,angles,quotes}
\colorlet{linecol}{black!75}
\usepackage{forest}
\tikzset{
  my rounded corners/.append style={rounded corners=2pt},
}

\usepackage{xcolor}

\biboptions{sort&compress}

\journal{Pattern Recognition}

\makeatletter
\def\ps@pprintTitle{%
 \let\@oddhead\@empty
 \let\@evenhead\@empty
 \def\@oddfoot{}%
 \let\@evenfoot\@oddfoot}
\makeatother

\begin{document}
\begin{frontmatter}

\title{
    Towards Automatic Threat Detection: A Survey of Advances of Deep Learning within X-ray Security Imaging
}


\author[label1,label2]{Samet Akcay}
\author[label2]{Toby Breckon}
\address[label1]{
    Intel R\&D, UK
}
\address[label2]{
    Department of Computer Science, Durham University, Durham, UK
}

\begin{abstract}
    X-ray security screening is widely used to maintain aviation/transport security, and its significance poses a particular interest in automated screening systems. This paper aims to review computerised X-ray security imaging algorithms by taxonomising the field into conventional machine learning and contemporary deep learning applications. The first part briefly discusses the classical machine learning approaches utilised within X-ray security imaging, while the latter part thoroughly investigates the use of modern deep learning algorithms.  The proposed taxonomy sub-categorises the use of deep learning approaches into supervised and unsupervised learning, with a  particular focus on object classification, detection, segmentation and anomaly detection tasks. The paper further explores well-established X-ray datasets and provides a performance benchmark.  Based on the current and future trends in deep learning, the paper finally presents a discussion and future directions for X-ray security imagery.
\end{abstract}


\end{frontmatter}



\section{Introduction}
\label{sec: introduction}

X-ray security screening is one of the most widely used security measures for maintaining airport and transport security, whereby manual screening by human operators plays the vital role. Although experience and knowledge are the key factors for confident detection, external variables such as emotional exhaustion and job satisfaction adversely impact the manual screening \cite{Chavaillaz2019}. 

Cluttered nature of X-ray bags also negatively affects the decision time and detection performance of the human operators \cite{Schwaninger2008, Wales2009}. For instance, the threat detection performance of human screeners significantly reduces when laptops are left inside the bags. This is due to the compact structure of laptops, limiting detection capability of the screeners \cite{Mendes2013}.  All these issues necessitate the use of automated object detection algorithms within X-ray security imaging, which would  maintain the alertness and improve detection and response time of human operators \cite{Chavaillaz2018}.

\begin{figure*}
    \centering
    \includegraphics[width=\linewidth]{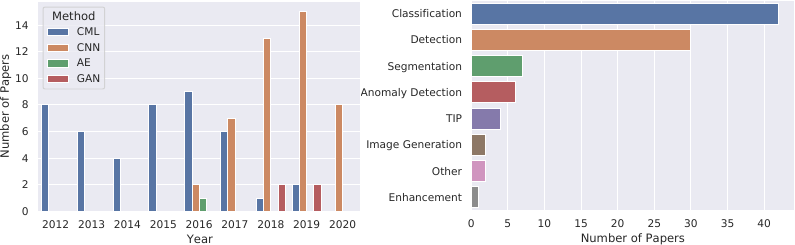}
    \caption{Statistics for the recent papers published in X-ray security imaging. (a) Distribution of machine learning vs. deep learning papers over the years. (b) Distribution of the papers based on the task}
    \label{fig:stats}
\end{figure*}

Despite the surge of interest in X-ray screening \cite{Murray1995, Zentai2008, Wells2012, Caygill2012, Singh2003}, automated computer-aided screening is understudied, particularly due to the lack of data, and the need for advanced learning algorithms. Previous work in the field have focused more on conventional image analysis \cite{Abidi2005, Abidi2006, Lu2006} and machine learning methods, spanning classification  \cite{Rogers2015,Kundegorski2016,Mery2016}, detection \cite{Franzel2012,Bastan2015} and segmentation \cite{Heitz2010, Kechagias-Stamatis2017} tasks. Notable surveys within the field \cite{Mouton2015, Rogers2016a}  thoroughly review these approaches and categorize the existing literature within image processing and understanding. 

More recently, on the other hand, deep-learning-based algorithms have been adopted in X-ray security imaging \cite{Akcay2016, Mery2017c, Jaccard2016b}, especially after convolutional neural networks (CNN) significantly outperform the conventional machine learning methods.  To this end, as of 2017, the use of deep learning algorithms is in the official US Government technology road-map for use across the US; and as of 2019/20, several early commercial systems have emerged from the academic research \cite{dhs2018}.

\begin{figure*}
    \centering
    \begin{tikzpicture}[
        level distance=1.25in,
        sibling distance=.25in,
        scale=.75
    ]
    \tikzset{edge from parent/.style= 
                {thick, draw,
                    edge from parent fork right},every tree node/.style={draw,minimum width=1in,text width=1in, align=center},grow'=right}
    \Tree 
        [. {X-ray Security Imaging}  
            [.{Conventional Image Analysis}
                [.{Image Enhancement} 
                    {\cite{Abidi2004, Abidi2005, Singh2005, Abidi2005a, Rogers2014, Rogers2017b}}
                ]
                [.{Threat Image Projection (TIP)} 
                    {\cite{Mitckes2003, Rogers2016, Mery2017b}}
                ]
            ]        
            [.{Machine Learning Algorithms}
                [.{Classification} 
                    [.{Single View}
                      \cite{Oertel2006,Gesick2009, Fu2009, Bastan2011,Turcsany2013,Bastan2013,Zheng2013,Zhang2014,Jaccard2014,Kolkoori2014,Zhang2014, Jaccard2014,Bastan2015,Rogers2015,Zhang2015,Zhang2015a, Rogers2015,Kundegorski2016,Mery2016}
                    ]
                    [.{Multi-View}
                        \cite{Abusaeeda2011,Zheng2013,Mery2011,Mery2012a,Mery2013b,Mery2013c,Mery2013a,Mery2015a,Mery2016a,Riffo2016,Mery2017c, Canizares2018}
                    ]
                ]
                [.{Detection} 
                    [.{Single View}
                        {\cite{Schmidt-Hackenberg2012, Bastan2015}}
                    ]
                    [.{Multi-View}
                        {\cite{Franzel2012,Bastan2015}}
                    ]
                ] 
                [.{Segmentation} 
                    {\cite{Paranjape1998, Sluser1999, Singh2004, Ding2006, Lu2006, Heitz2010, Kechagias-Stamatis2017}}
                ]                
            ]
            [.{Deep Learning Algorithms}
                [.{Supervised}
                    [.{Classification} 
                        [.{Single View}
                            {\cite{Akcay2016,SvecP.2016, Jaccard2016b, Jaccard2016, Jaccard2017,Jaccard2016a,Rogers2017a,Caldwell2017, Yuan2018,Zhao2018, Xu2018,Miao2019}}
                        ]
                    ]
                    [.{Detection} 
                        [.{Single View}
                            [.{Region-based}
                                {\cite{Akcay2017, Akcay2018a, Hassan2019}}
                            ]
                            [.{Single Shot}
                                {\cite{Akcay2018a,Liu2018,Xu2018, Cui2019a, Hassan2019, Subramani2020, Hassan2020}}
                            ]
                        ]
                        [.{Multi-View}
                            [.{Region-based}
                                {\cite{Steitz2018, Liang2019, Isaac-Medina2020}}
                            ]
                        ]
                    ]
                    [.{Segmentation}
                        {\cite{Gaus2019a, Gaus2019b, Hassan2020b}}
                    ]
                ]
                [.{Unsupervised}
                    [.{Anomaly Detection}
                        \cite{Tuszynski2013,Andrews2016,Andrews2016a, Andrews2017,Akcay2018b,Akcay2019,Griffin2019}
                    ]
                ]
            ] 
        ]
    \end{tikzpicture}      
    
    \caption{A Taxonomy of the X-ray security imaging papers.}
    \label{fig:taxonomy}
\end{figure*}
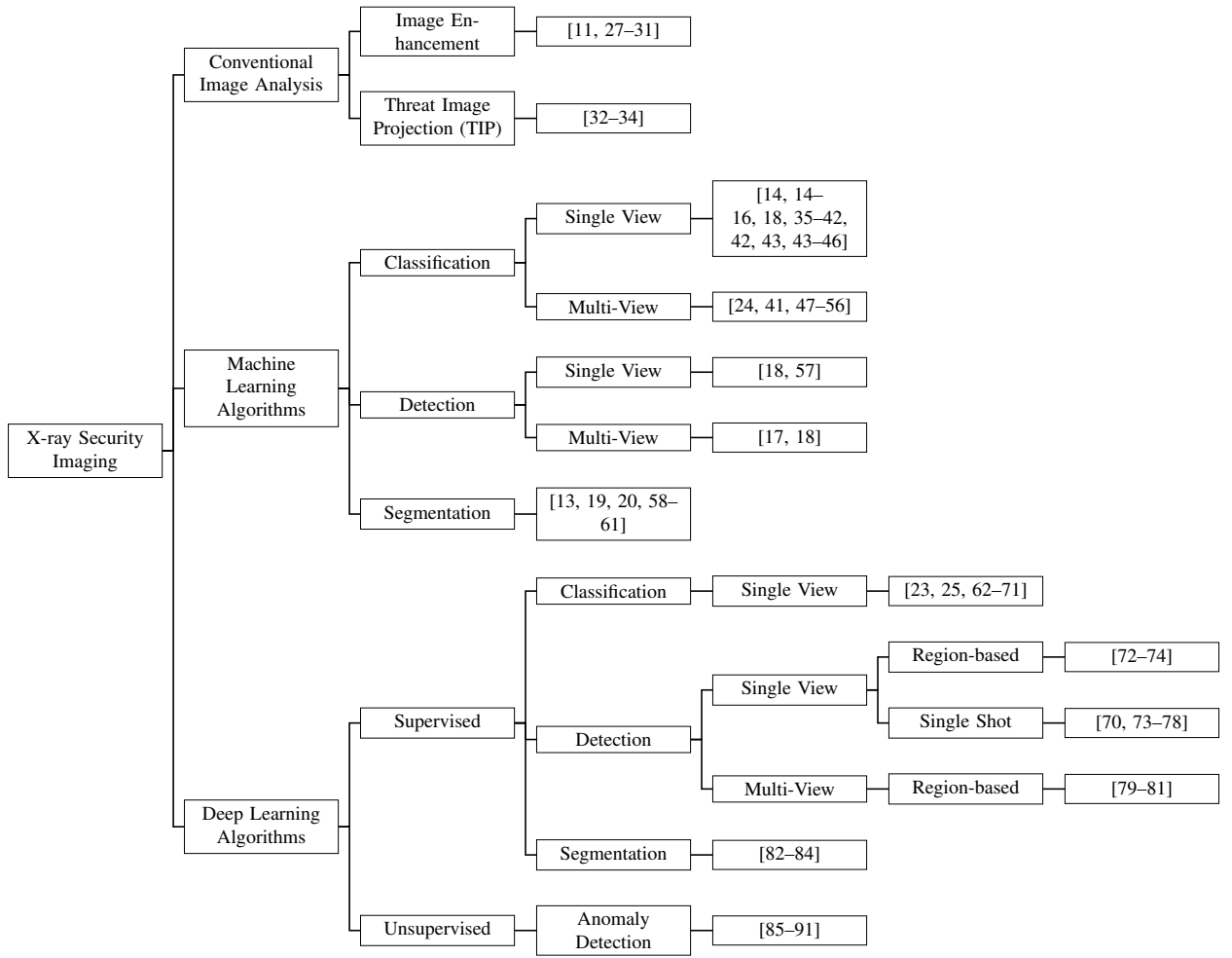

Following this trend change, this literature survey reviews the published work within various computer vision tasks (Figure \ref{fig:stats}B) in X-ray security screening, with a particular focus on the deep learning applications. We use the following keywords and operators in Google Scholar search to search the relevant papers: `((x-ray security) OR (x-ray baggage) OR (x-ray luggage)) AND ((detection) OR classification OR segmentation)'. We also conduct a backward search based on the citations and related papers, and overall identify approximately 213 relevant articles, of which 36 employ deep-learning-based algorithms (Figure \ref{fig:stats}A). Based on the scope of the work, we finally reduce the number of relevant papers to 130. The main contributions of this work, therefore, are as follows:

\begin{itemize}
    \item \emph{taxonomy} ---  an extensive overview of classical machine learning and contemporary deep learning within X-ray security imaging (Figure \ref{fig:taxonomy}).
    \item \emph{datasets} --- an overview of the large datasets used to train deep learning approaches within the field.
    \item \emph{open problems} --- discussion of the open problems, current challenges, and future directions based on the current trends within computer vision.
\end{itemize}

The rest of the paper is as follows: Section \ref{sec:background} provides a brief background regarding the principle of X-ray imaging. Sections \ref{sec:datasets} and \ref{sec:evaluation-criteria} introduce datasets and evaluation criterion used to measure the performance of the methods. Sections \ref{sec: conventional-image-analysis} and \ref{sec:ml} explore conventional image analysis and machine learning algorithms. Section \ref{sec:dl.applications} reviews the applications of the deep learning algorithms within X-ray security imaging. Section \ref{sec:discussion} discusses the open problems, current challenges and  Section \ref{sec:conclusion} finally concludes the paper. 
\section{Background: X-ray Imaging}
\label{sec:background}

\begin{figure}
    \centering
    \includegraphics[width=0.75\linewidth]{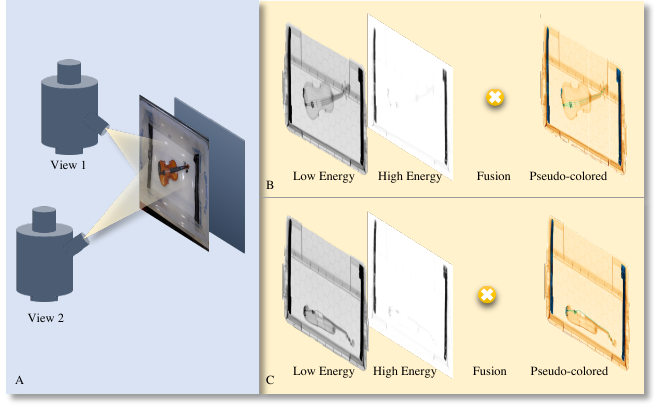}
    \caption{High-level overview of X-ray imaging. RGB and X-ray images are from COMPASS-XP dataset \cite{Caldwell2019}.}
    \label{fig:xray-imaging}
\end{figure}

As depicted in Figure \ref{fig:xray-imaging}A, the main principle of X-ray imaging is that an X-ray tube generates beams that penetrate the scanned object. Depending on its material density, the object attenuates the X-ray signal. This attenuation is formulated as $I_x = I_0e^{\mu x}$, where $I_x$ is the intensity at $x$ cm, $I_0$ is the initial intensity, and $\mu$ is the linear attenuation coefficient based on the thickness of the material. This formulation shows that material density and measured intensity are inversely proportional —for instance, a high-density material yields high attenuation and low measured intensity.

Modern X-ray machines are equipped with multiple ($m$)-energy that produces $m$ X-ray images via different energies (Figure \ref{fig:xray-imaging}B), identifying the objects' density and effective atomic number ($Z_{eff}$). The estimated intensity and $Z_{eff}$ values are converted to pseudo-colored images via a look-up table \cite{Abidi2005a}. In addition to multiple energy levels, the state-of-the art machines generates X-ray scans from multiple view-points to view the objects of interest from various angles (Figure \ref{fig:xray-imaging}C). For more details regarding the X-ray image generation process, the reader is referred to \cite{Mery2015Computer}.
\section{Datasets}
\label{sec:datasets}
This section explores X-ray security imaging datasets that are widely used in the literature.

\subsection{Durham Baggage (DB) Patch/Full Image Dataset}
\label{datasets.db}
This dataset comprises $15,449$ X-ray samples with associated false color  materials  mapping  from  dual-energy four-view Smiths 6040i machine. Originally, samples have the following class distributions: $494$ \textit{camera}, $1,596$ \textit{ceramic knife}, $3,208$ \textit{knife}, $3,192$ \textit{firearms}, $1,203$ \textit{firearm parts}, $2,390$ \textit{laptop} and $3,366$ \textit{benign} images. Several variants of this dataset is constructed for classification (DBP2 and DBP6) \cite{Akcay2016, Kundegorski2016, Akcay2018a} and detection (DBF2 and DBF6) \cite{Akcay2017, Akcay2018a}. The dataset is well-balanced with wide variety of threat objects. However, being a private dataset, it's usage is limited within the literature.



\subsection{GDXray}
\label{datasets.gdxray}
Grima X-ray Dataset ($\mathbb{GDX}$RAY) \cite{Mery2015} comprises $19,407$ X-ray samples from five various subsets including castings  ($2,727$), welds ($88$), baggage ($8,150$), natural images ($8,290$), and settings ($152$). 
The baggage subset is mainly used for security applications and comprises images from multiple-views. The limitation of this dataset is its non-complex content, which is non-ideal to train for real-time deployment.

\subsection{UCL TIP}
\label{datasets.ucl}
This dataset comprises $120,000$ benign images, scanned with Rapiscan\textsuperscript{®} R60. Each sample is 16-bit grayscale with sizes varying between $1920 \times 850$ and $2570 \times 850$. The train and test split of the dataset is $110000$ : $10000$, where the training images are $256 \times 256$ patches randomly sub-sampled from $110,000$ images and the test set comprises $5000$ benign and $5000$ threat images. The threat images are synthetically generated via the TIP algorithm proposed in \cite{Rogers2016}, where, depending on the application, small metallic threats (SMT) or car images are projected into the benign samples. With several variants, this dataset is used in several studies such as \cite{Andrews2016a, Andrews2017, Jaccard2016, Jaccard2016a, Jaccard2016b, Jaccard2017, Rogers2017a}.

\subsection{SIXray}
\label{datasets.sixray}
With unknown machine specification, this dataset is acquired from subway stations and released by \cite{Miao2019}, SIXray dataset comprises $1,059,231$ X-ray images, $8929$ of which are manually annotated for $6$ different classes: gun, knife, wrench, pliers, scissors, hammer, and background. The dataset consists of objects with a wide variety in scale, viewpoint and mostly overlapping, making it a suitable dataset for real-time classification, detection and segmentation applications.

\subsection{Durham Baggage Anomaly Dataset --DBA}
\label{datasets.uba}
This in-house dataset comprises $230,275$ dual energy X-ray security image patches extracted via a $64 \times 64$ overlapping sliding window approach. The dataset contains 3 abnormal sub-classes —\textit{knife} ($63,496$), \textit{gun} ($45,855$) and \textit{gun component} ($13,452$). Normal class comprises $107,472$ benign X-ray patches, split via $80:20$ train-test ratio. DBA dataset is used in \cite{Akcay2018b} and \cite{Akcay2019} for unsupervised anomaly detection. Similar to DB dataset varians, this dataset is not publicly available, limiting its use in the literature.

\subsection{Full firearm vs Operational Benign --FFOB}
\label{datasets.ffob}
As presented in \cite{Akcay2018a, Akcay2018b, Akcay2019, Griffin2019}, this dataset contains samples from  the UK government evaluation dataset \cite{CAST2016}, comprising both expertly concealed firearm (threat) items and operational benign (non-threat) imagery from commercial X-ray security screening operations (baggage/parcels). Denoted as FFOB, this dataset comprises $4,680$ firearm full-weapons as full abnormal and $67,672$ operational benign as full normal images, respectively. The main drawback of this dataset is its restricted access.

\subsection{Compass - XP Dataset}
\label{datasets.compass-xp}
This dataset \cite{Caldwell2019} is collected using $501$ objects from $369$ object classes that are a subset of ImageNet classes. The dataset includes $1901$ image pairs such that each pair has an X-ray image scanned with Gilardoni FEP ME 536 and its photographic version is taken with a Sony DSC-W800 digital camera. Besides, each X-ray image has its low-energy, high-energy, material density, grey-scale (the combination of low and high energy) and pseudo-coloured RGB versions. This dataset is well-suited to X-ray imaging research; however, its non-cluttered nature limits its use for real-time applications.

\subsection{OPIXray Dataset}
\label{datasets.opixray}
OPIXray dataset \cite{Wei2020} is an airport inspection dataset manually annotated by the security personnel. The dataset comprises 8885 X-ray images (7019 training, 1776 testing) from five sharp objects, including folding knife (1,993), straight knife (1,044), scissor (1,863), utility knife (1,978) and multi-tool knife (2,042).

\begin{table*}
\centering
\resizebox{\textwidth}{!}{%
\begin{tabular}{@{}lllllll@{}}
\toprule
Dataset     & Domain  & Task              & \# Samples & Classes                              & Performance     & Reference \\ \midrule
DBP2        & Baggage & Classification    & 19,938     & firearm, background                  & ACC: 0.994      & \cite{Akcay2016, Akcay2018a}          \\
DBP6        & Baggage & Classification    & 10,137     & firearm, firearm parts, camera,      & ACC: 0.937      & \cite{Akcay2016, Akcay2018a}           \\
            &         &                   &            & knife, ceramic knife, laptop         &                 &           \\
UCL TIP     & Cargo   & Classification    & 120,000    & small metallic threat (SMT), car     & ACC: 0.970      & \cite{Jaccard2016a, Andrews2016a, Caldwell2017, Rogers2017a, Andrews2017, Jaccard2017}          \\
            &         & Detection         &            &                                      &                 &           \\
            &         & Anomaly Detection &            &                                      &                 &           \\
GDXRay      & Baggage & Classification    & 19,407     & gun, shuriken, razor blade           & ACC: 0.963      & \cite{Mery2017c, Mery2017a, Xu2018, Sangwan2019}          \\
            &         & Detection         &            &                                      &                 &       \\
DBF2        & Baggage & Detection         & 15,449     & firearm, background                  & mAP: 0.974      & \cite{Akcay2017, Akcay2018a}          \\
DBF6        & Baggage & Detection         & 15,449     & firearm, firearm parts, camera,      & mAP: 0.885      & \cite{Akcay2017, Akcay2018a}            \\
            &         &                   &            & knife, ceramic knife, laptop         &                 &           \\
PBOD        & Baggage & Classification    & 9,520      & Explosives                           & AUC: 0.950      & \cite{Morris2018}          \\
MV-Xray     & Baggage & Detection         & 16,724     & Glass Bottle, TIP Weapon, Real Weapon& mAP: 0.956      & \cite{Steitz2018}          \\
SASC        & Baggage & Detection         & 3,250      & Scissors, Aerosols                   & mAP: 0.945      & \cite{Liu2018}           \\
Zhao \etal  & Baggage & Classification    & 1,600      & wrench, pliers, blade, lighter,      & ACC: 0.992      & \cite{Zhao2018}           \\
            &         &                   &            & knife, screwdriver, hammer           &                 &           \\
Smiths-Duke & Baggage & Detection         & 16,312     & gun, pocket knife, mixed sharp       & mAP: 0.938      & \cite{Liang2018}          \\
SIXray      & Baggage & Detection         & 1,059,231  & gun, knife, wrench, pliers,          & mAP: 0.439      & \cite{Miao2019}           \\
            &         &                   &            & scissors, hammer, background         &                 &            \\
UBA         & Baggage & Anomaly Detection & 230,275    & gun, gun part, knife                 & AUC: 0.940      & \cite{Akcay2018b, Akcay2019}          \\
FFOB        & Baggage & Anomaly Detection & 72,352     & full-weapon, benign                  & ACC: 0.998      & \cite{Akcay2018b, Akcay2019}          \\
Yang \etal  & Baggage & Classification    & 2,000      & wrench, fork, handgun, power bank,   & ACC: 0.991      & \cite{Yang2019}          \\
            &         &                   &            & lighter, pliers, knife, liquid, umbrella, screwdriver &&           \\ 
OPIXray  & Baggage & Detection    & 8,885      & Folding Knife, Straight Knife,   & mAP: 0.753      & \cite{Wei2020}          \\
            &         &                   &            & lighter, Scissor, Utility Knife, Multi-tool Knife &&           \\ 
\bottomrule
\end{tabular}%
}
\caption{Datasets used in deep learning applications within X-ray security imaging}
\label{tab:datasets}
\end{table*}




\section{Evaluation Criteria}
\label{sec:evaluation-criteria}
Before reviewing the papers, it is essential to introduce the various performance metrics used in the field. 
All of the metrics shown here are computed based on true positives ($TP$), false positives ($FP$), true negatives ($TN$) and false negatives ($FN$).

\paragraph{Accuracy} (ACC) is defined as the number of correctly predicted samples over the the total number of predictions, which is mathematically shown as $ACC = (TP+TN)/(TP+TN+FP+FN)$.

\paragraph{True Positive Rate} (TPR) is the proportion of correctly predicted positive samples: $TPR = TP / (TP + FN)$

\paragraph{False Positive Rate} (FPR) is calculated as the ratio of the negative samples predicted as positive: $FPR = FP / (FP + TN)$.

\paragraph{Mean Average Precision} (mAP) is defined as the mean of the average precision, a metric evaluated by the area under the precision and recall curve, where precision is $TP/(TP+FP)$, and recall is $TP/(FN+TP)$.

\paragraph{Area Under the Curve}
(AUC) is the area under the curve of the receiver operating characteristics (ROC), plotted by the true positive rates and false positives rates.

Table \ref{tab:datasets} shows the benchmark statistics based on the datasets and evaluation criteria discussed in Sections \ref{sec:datasets} and \ref{sec:evaluation-criteria}. The best performing models will be explained in the following sections.
\section{Conventional Image Analysis}
\label{sec: conventional-image-analysis}

This section explores the conventional image analysis techniques that perform image enhancement and threat image projection. 

\subsection{Image Enhancement}


Preprocessing the input data plays a substantial role to yield higher-quality images that increase the readability by both screener and computer. Common approach in literature is to \cite{Abidi2005} fuse low and high energy X-ray images and apply background subtraction for noise reduction, followed by either manual \cite{Abidi2004} or adaptive \cite{Singh2005} threshold selection. Pseudo-colouring  \cite{Abidi2005, Abidi2005a, Chan2010} is another enhancement technique that colours grey scale X-ray images, improving the detection performance and alertness level of the operators.



\subsection{Threat Image Projection } 
\label{subsec:ml.tip}
Threat image projection (TIP) \cite{Mitckes2003} is another method that could be categorised within conventional image analysis. TIP is used to generate a synthetic dataset to either train human screeners \cite{Cutler2009} or machine/deep learning models. A common TIP approach is to project a binary threat mask onto a benign input X-ray image via multiplication, yielding an output X-ray image with the threat item. Application of affine  \cite{Rogers2016} or logarithmic \cite{Mery2017b} transformations adds various threat projections onto the benign image. Empirical studies show that the use of TIP improves the overall detection performance of models \cite{Rogers2016, Mery2017b, Bhowmik2019}.



\section{Machine Learning Approaches in X-ray Security Imaging}
\label{sec:ml}


This section explores the applications of  conventional machine learning approaches in X-ray security imaging. The literature is reviewed based on three tasks:  classification,  detection, and  segmentation. For an alternative perspective for this section, the reader could refer to the related reviews of Mery \cite{Mery2015Computer} and Rogers \etal \cite{Rogers2016a}.

\subsection{Object Classification}
\label{sec:ml.classification}


Prior to the dominance of the deep learning within the field, the bag of visual words (BoVW) approach was prevalent. A common approach is to (i) perform feature extraction via detector/descriptors, (ii) cluster the features via k-means \cite{Hartigan1979, Bastan2011} and (iii) classify RF\cite{Breiman2001}, SVM \cite{Hearst1998} or sparse-representation \cite{Bastan2011, Mery2012a, Turcsany2013, Jaccard2014, Zhang2014, Zhang2015, Zhang2015a, Mery2016, Kundegorski2016}.

Despite the BoVW dominance, other computer vision/machine learning techniques have also been studied for X-ray object classification task. 
Mery \etal \cite{Mery2012a} utilize structure estimation and segmentation together with a general tracking algorithm to detect X-ray objects. Similar works \cite{Zheng2013, Mery2016a, Riffo2016, SvecP.2016, Mery2017c, Xu2019} exhaustively evaluate various computer vision techniques, with a specific focus on k-NN based sparse representation, achieving comparable accuracy to deep models on \hyperref[datasets.gdxray]{GDXray} ($94.7 \%$ vs. $96.3\%$). 


\subsection{Object Detection}
\label{ml.detection}

This section reviews the conventional X-ray object detection models presented in the literature. Being a challenging task, where the bounding box coordinates and class labels are to be predicted simultaneously, conventional object detection algorithms in the literature is relatively limited in the field.


Similar to classification methods explored in Section \ref{sec:ml.classification}, conventional detection algorithms also primarily employ BoVW approach. Evaluation works of \cite{Schmidt-Hackenberg2012, Bastan2013, Bastan2015} exhaustively investigate the use of BoVW for the X-ray object detection. Evaluating various feature descriptors with  SVM classifier \cite{Hearst1998} shows that (i) sparse intensity domain image descriptor (SPIN) \cite{Lazebnik2005} achieves the highest detection performance (mAP: $46.1 \%$).

Unlike classification, here the models also utilise multiple-view imagery, which generally improves the performance when rotation and superimposition hinder the viewability of the objects from one view \cite{Michel2009}. Despite its computational complexity, multi-view imaging help human operators and machines to improve the detection performance \cite{Bastian2008, Franzel2012, Bastan2015}. A general multi-staged approach proposed in the works of \cite{Mery2017, Mery2013b, Mery2013a, Mery2011} initially performs feature extraction via feature descriptors and k-NN classifier \cite{Cover1967}. Features matched from different views are classified by the k-NN classifier \cite{Cover1967} ($95.7\%$ precision).

\subsection{Object Segmentation}
\label{sec:ml.segmentation}


This section explores various segmentation techniques presented in the literature. Early work in the field \cite{Paranjape1998, Sluser1999} investigates simplistic pixel-based segmentation with a fixed absolute threshold and region grouping. Subsequent work, on the other hand, focuses more on pre-segmentation via nearest neighbour, overlapping background removal and final classification \cite{Singh2004, Ding2006, Lu2006, Heitz2010, Kechagias-Stamatis2017}.

Another approach is to utilize graph-based algorithms for the segmentation. Early work concentrates on fuzzy similarity distance between attribute relational graphs \cite{Wang2005, Ding2006}, while more recent investigates spectral clustering and variational image segmentation \cite{Mallia-Parfitt2019}.


Despite the promising detection performance reported, these techniques are generally experimented on a small datasets, limiting their scalability for real-time applications. 




\section{Deep Learning in X-ray Security Imaging}
\label{sec:dl.applications}
This section reviews the X-ray security applications utilising deep learning algorithms. As shown in Figure \ref{fig:taxonomy} and Table \ref{tab:overview.of.dl.applications}, we categorise the algorithms as supervised (classification, detection and segmentation) and unsupervised (anomaly detection) approaches.

\subsection{Supervised Approaches}
Supervised approaches are grouped within classification, detection and segmentation tasks, where the models utilise the ground-truth global, bounding-box and pixel-wise labels, respectively.

\begin{figure}
    \centering
    \includegraphics[width=\linewidth]{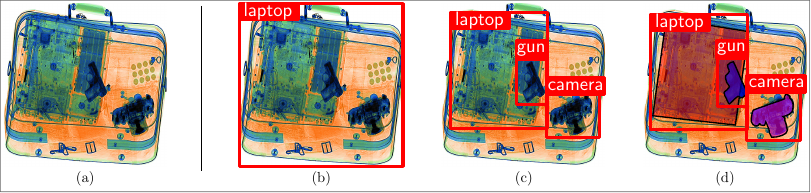}
    \caption{An input X-ray image, and the outputs depending on the deep learning task, (a) classification via ResNet-50 \cite{He2015}, (b) detection with YOLOv3 \cite{Redmon2018} and segmentation via Mask RCNN \cite{He2017}}
    \label{fig:dl}
\end{figure}

\subsubsection{Classification}
\label{subsec: applications.classification}
The study of \cite{Akcay2016} is one of the first research applying CNN to X-ray security imagery as a classification task, where the model predicts the global image label (Figure \ref{fig:dl}B). The authors examine the use of CNN via transfer learning to evaluate to what extent transfer learning helps classify X-ray objects within the problem domain, where the availability of the datasets is somewhat limited. Freezing AlexNet weights layer by layer on a two-class (\textit{gun vs. no-gun}) X-ray classification problem shows that CNN significantly outperforms the BoVW approach (SIFT+SURF), trained with SVM or RF, even when the layers of the network are all frozen. Another set of experimentation analyses the use of CNN within a challenging 6-class classification problem, whose results show a great promise of the use of CNN in the field. 

A similar work \cite{Jaccard2016b} compares the use of deep learning against the conventional machine learning to classify non-empty cargo containers with cars or SMT. A multi-stage approach first classifies cargo containers as empty vs. non-empty. The second stage is the classification of cars from the containers classified as non-empty, achieved via oBIF + RF. By using \hyperref[datasets.ucl]{UCL TIP} dataset, the authors evaluate the performance of 9 and 19-layer networks \cite{Jaccard2016} that are similar to \cite{Krizhevsky2012} and \cite{Simonyan2014}, and show that even the worst-performing CNN outperforms the conventional machine learning approach (oBIF + RF).

A follow-up work \cite{Jaccard2017} further investigates the detection of cars from X-ray cargo images. A sliding window splits \hyperref[datasets.ucl]{UCL TIP} images into patches. Authors then explore various features including intensity, oBIF \cite{Griffin2009}, Pyramid of Histogram of Visual Words (PHOW) \cite{Bosch2007} and CNN features. Training these features on SVM \cite{Hearst1998}, RF \cite{Breiman2001}, and soft-max (CNN) shows that a RF classifier trained on the VGG-18 \cite{Simonyan2014} features extracted from log-transform images achieves the highest performance (FPR: $0.22$\%). 

Additional work by Jaccard \etal \cite{Jaccard2016a} evaluate the impact of input types on CNN performance by training single-channel raw image and dual-channel data that contains the raw image and its log-transformed image on VGG \cite{Simonyan2014} variants. The quantitive analysis demonstrates that VGG-19 model trained from scratch by using dual-channel raw and log-transformed images outperforms the other variants (AUC: $97\%$, FPR: $6\%$).

Rogers \etal \cite{Rogers2017a} explore the use of dual-energy X-ray images for automated threat detection. Authors investigate varying transformations applied to high-energy ($H$) and low-energy($L$) X-ray images captured via the dual-energy X-ray machine. Using \hyperref[datasets.ucl]{UCL TIP} dataset,  640,000 image patches are generated via a $256 \times 256$ sliding-window. Training this dataset with a fixed VGG-19 network \cite{Simonyan2014} with varying input channels, including single-channel ($H$), dual-channel($\{H, -\log{H}\}$,  $\{ -\log{H}, -\log{L}\}$) and four-channels 
$(\{ -\log{L}, L, H, \allowbreak -\log{H} \})$
shows that dual and four-channels always achieves superior detection performance compared to their single-channel variants (ACC: $95\%$--dual vs $90\%$--single).

Inspired by the limited availability of X-ray datasets, a  three-stage algorithm by Zhao \etal \cite{Zhao2018} initially classifies and labels the input X-ray dataset via the angle information of the foreground objects extracted from the input image. The second stage generates new X-ray objects via an adversarial network similar to \cite{Arjovsky2017}. Additional use of \cite{Isola2016} improves the quality of the generated images. Finally, a small classification network confirms whether the generated image belongs to the correct class. In a follow-up study, Yang \etal \cite{Yang2019} further investigate the ways to improve the GAN training to produce better X-ray images. The quantitative evaluation shows that the proposed GAN approach in the paper generates visually superior prohibited items.

Miao \etal \cite{Miao2019} introduce a model (CHR) to classify/detect X-ray images from \hyperref[datasets.sixray]{SIXray}. The model copes with class imbalance and clutter issue by extracting image features from three consecutive layers, where subsequent layers are upsampled and concatenated with the previous layers. A refinement function $g()$ removes the redundant information from the concatenated feature map. The objective of the work is to minimize the loss of the weighted sum of the classification of the refined mid-level features from the three consecutive layers ($\{h(\tilde{x}_n^{(l-1)}), \allowbreak h(\tilde{x}_n^{(l)}), h(\tilde{x}_n^{(l+1)}) \}$). Training the model with the proposed loss yields $2.13\%$ mAP improvement when used with ResNet-101 on \hyperref[datasets.sixray]{SIXray} ($36.01$ vs. $38.14$). A similar approach \cite{Wei2020} introduces a plug and play module that utilises edge and material information to localise objects via attention mechanism.

An evaluation work \cite{Morris2018} investigates the use of CNN for the task of explosive detection. An initial stage process the input data by fixing the image size, cropping the irrelevant background object where $Z_{eff}=0$ and applying data augmentation transformations. Evaluation of random initialization vs. pre-training on VGG19\cite{Simonyan2014}, Xception \cite{Chollet2017}, and InceptionV3 \cite{Szegedy2016} networks shows that randomly initialized models achieves superior accuracy for binary classification task. To study the impact of intensity and Z-eff values on the performance, the authors train three VGG-19 networks on both intensity and Z-effective, the intensity only and Z-effective only. Training the model with only Z-eff is shown to yield the highest accuracy. The final set of experiments investigates localization via heatmaps and shows that pre-trained networks achieves superior performance since randomly initialized networks tend to overfit on small datasets.

Caldwell \etal \cite{Caldwell2017} study the generalisation capability of models trained with different datasets from various scanners. The authors create training and test splits from both single or multiple domains to investigate the impact of transferring between other modalities. Quantitative analysis reveals that transferring information is challenging due to unknown parameters of the scanners and generalisability of CNN to the unseen target dataset.

\subsubsection{Detection}
\label{subsec: applications.detections}
This section explores CNN-based object detection algorithms by a categorisation of single and multi-view object detection.

\paragraph{Single-View Detection}

After the success of CNN for classification, the work of \cite{Akcay2017} trains sliding-window based CNN, Faster RCNN \cite{Ren2015} and R-FCN \cite{Dai2016b} models on \hyperref[datasets.db]{DBF2/6} datasets for firearm and multi-class detection problems. Experiments demonstrate that Faster RCNN \cite{Ren2015} with VGG16 \cite{Simonyan2014} yield $88.3\%$ mAP on 6-class \hyperref[datasets.db]{DBF6} dataset, while R-FCN with ResNet-101 achieves the highest performance ($96.3$ mAP) on 2-class (\textit{gun} vs \textit{no-gun}) on \hyperref[datasets.db]{DBF2} dataset.

Sigman \etal \cite{Sigman2020} utilise an adversarial domain adaptation technique to match the distribution of the background of a sizeable unlabelled stream of commerce (SoC) dataset. By doing so helps to detect the objects in the SoC dataset by training a Faster RCNN \cite{Ren2015} on a small labelled dataset.

Subramani \etal \cite{Subramani2020} investigate the use of SSD \cite{Liu2016} and RetinaNet \cite{Lin2018} trained on SIXray10 dataset, achieving $60.5\%$ and $60.9\%$, respectively.

Liu \etal \cite{Liu2018} also performs object detection via YOLOv2 \cite{Redmon2018} to detect scissors and aeorosols on SASC dataset. Training YOLO v2 for 6000 iterations yield $94.5 \%$ average precision and $92.6 \%$ recall rates with $68$ FPS run-time speed.

Cui and Oztan \cite{Cui2019a} argue that RetinaNet \cite{Lin2018} achieves comparable detection performance, while being considerably faster than traditional sliding window classification when trained with 30,000 images synthetically generated via TIP with $5000$ X-ray cargo containers and $544$ firearms. 

Hassan \etal \cite{Hassan2019} proposes an object detection algorithm, whereby the RoI is generated via cascaded multi-scale structure tensors that extracts based on the variations of the orientations of the object. The extracted RoI is then passed into a CNN, which quantitatively and computationally outperforms RetinaNet, YOLOv2 and F-RCNN on \hyperref[datasets.gdxray]{GDXray} and \hyperref[datasets.sixray]{SIXray} datasets. A similar approach in \cite{Hassan2020, Hassan2020a} produces contour-based object proposals, which are subsequently forward-passed into a CNN, achieving $96\%$ mAP on SIXray10 dataset.

Motivated by the lack of annotated X-ray datasets, Xu \etal \cite{Xu2018},  make use of attention mechanisms for the localization of threat materials. The first stage forward-passes an input and finds the corresponding class probability. The back-propagation step identifies the interconnected neurons activated during the decision of the output class.  Activations from the first convolutional layer generate a heatmap. The final stage refines the activation map by normalizing the layers with the activations of the previous layer. Comparison against the traditional deconvolution method (mAP: $34.3\%$) shows that the proposed method achieves superior detection ($56.6\%$) without requiring bounding box information.

Similar to \cite{Caldwell2017}, generalisation capability of CNN is studied by Gaus \etal \cite{Gaus2019a} by training/validating CNN on different datasets (DBF3 ($88 \%$ mAP) $\rightarrow$ SIXray ($85\%$ mAP)).

\paragraph{Multi-View Detection}
There are number of papers utilising the multi-view X-ray imagery to improve the detection performance of their models. An evaluation work \cite{Liang2018} explores the performance of  F-RCNN, R-FCN \cite{Dai2016b} and SSD \cite{Liu2016} within single/multi-view X-ray imagery. Utilizing \textit{OR-gate} detection by merging object detection outputs from individual views shows that multi-view outperforms that of single-view ($0.938$ vs. $0.798$ when trained with R-FCN and ResNet-101). A two-stage approach by Liu \etal \cite{Jinyi2019} first extracts foreground objects and subsequently utilises F-RCNN to detect $32,253$ subway X-ray images, with an mAP of $77\%$ for 6 object classes. 

A similar study \cite{Liang2019} explores SSD and F-RCNN by training on a dataset containing 4 threat classes, each of which comprises approximately $3,400$ images. F-RCNN with Inception ResNet v2 backbone yields the highest mAP ($92.2$ and $97.7$ on single and multi-view images, respectively). Another work \cite{Steitz2018} utilize multi-view by modifying F-RCNN. A multi-view pooling layer constructs 3D feature 2D extracted from the convolutional layers. 3D region proposal network generates the RoI. Classification and bounding box prediction is performed after 3D RoI pooling layer. Experiments show that multi-view yields an improvement compared to single-view imagery ($95.56 \%$ vs. $91.23 \%$). 

Isaac-Medina \etal \cite{Isaac-Medina2020} train a YOLOv3 \cite{Redmon2018} detector by utilising epipolar constraints of multiple-views of X-ray images, which outperforms the single-view by 2.2\% (Figure \ref{fig:dl}C).

Overall, these results suggest that the use of multiple-view imagery aids improving the detection performance of the deep learning models.


\subsubsection{Segmentation}
\label{subsec: applications.segmentation}
Due to the scarcity of datasets with pixel-level annotation, the task of segmentation is understudied within the field. One of the published work \cite{Gaus2019b} addresses segmentation and anomaly detection tasks together, whereby a dual-CNN pipeline initially segments RoI via Mask RCNN \cite{He2017} and classifies the regions as benign/abnormal via ResNet-18 \cite{He2015}, achieving $97.6\%$ segmentation mAP and $66.0\%$ anomaly detection accuracy (Figure \ref{fig:dl}D). Another work \cite{Bhowmik2019a} proposes three-stage approach, whereby (i) object-level segmentation is achieved by the use of Mask RCNN \cite{He2017}, (ii) sub-component regions are segmented via super-pixel segmentation and (iii) final object classification is performed via fine-grained CNN classification, which overall yields $97.91\%$ anomaly detection accuracy on $7,878$ electronic items. An \etal \cite{An2019} propose a segmentation model that utilises dual attention mechanism within an encoder-decoder segmentation network. The former attention module classifies the RoI, while the latter localises the object. Experiments on PASCAL alike structured X-ray dataset containing $7,532$ augmented images from 7-classes yield $99.3$ accuracy and $68.3$ mean intersection over union (mIoU).

Hassan \etal \cite{Hassan2020b} propose a single-stage instance segmentation algorithm. The method initially extracts transitional patterns via trainable structure tensors, which are subsequently passed to an encoder-decoder to construct the binary segmentation masks. mAP evaluation on GDXray ($96.7$), SIXray ($96.16$), OPIXray($75.32$) and COMPASS XP ($58.4$) datasets show that the model achieves the state-of-the-art instance segmentation performance on benchmarks.

\subsection{Unsupervised Approaches}
\label{subsec:dl.applications.anomaly.detection}
This section explores unsupervised deep learning models, where the proposed algorithms mainly investigate the anomaly detection task. Human operators tend to perform better detection when focusing on benign objects rather than threat items. Besides, the knowledge of every-day benign objects leads to a much better detection performance \cite{Sterchi2017}. The same concept is applied in anomaly detection, where the model is only trained with normal samples, and tested on normal/abnormal examples.

An anomaly detection approach \cite{Andrews2016} employs sparse feed-forward autoencoders in an unsupervised manner to learn the feature encoding of normal and abnormal data. An SVM \cite{Hearst1998} then classifies the images either anomalous or benign. Validation on MNIST \cite{Lecun1998} and freight container dataset (\textit{empty} vs \textit{non-empty}) shows that hidden layer representation extracted from the autoencoder, is significant for the detection of abnormalities in the images. When fused with the raw-input and residual error, features encoding from the hidden layers yield even better detection performance. 

A follow-up work utilizes intensity, log-intensity and VGG-19 \cite{Simonyan2014} features extracted from patches from \hyperref[datasets.ucl]{UCL TIP} dataset and train normal images via forest of random split trees anomaly detector \cite{Liu2012}. Testing the model on normal + abnormal data yields $64$\% AUC.

A similar study \cite{Akcay2018b}, in which image and latent vector spaces are optimized for anomaly detection, utilizes an adversarial network such that the generator comprises encoder-decoder-encoder sub-networks. The objective of the model is to minimize the distance between both real/generated images and their latent representations jointly, which overall outperforms the previous \sota  both statistically and computationally (\hyperref[datasets.uba]{UBA}: $64.3\%$, \hyperref[datasets.ffob]{FFOB}: $88.2\%$ -- AUC). A follow-up work \cite{Akcay2019} improves the performance of \cite{Akcay2018b} further by (i) utilizing skip-connections in the generator network to cope with higher resolution images, and (ii) learning the latent representations within the discriminator network (\hyperref[datasets.uba]{UBA}: $94.0\%$, \hyperref[datasets.ffob]{FFOB}: $90.3\%$ -- AUC).

\begin{landscape}
    \begin{table}
        \resizebox{\linewidth}{!}{%
        \begin{tabular}{@{}lllll@{}}
        \toprule
        Reference                                   & Domain         & Problem                                 & Method                           \\ \midrule
        Ak\c{c}ay \etal \cite{Akcay2016}            & Baggage        & Object Classification                   & CNN with transfer learning                              \\
        Svec \cite{SvecP.2016}                      & Baggage        & Object Classification                   & CNN with transfer learning                              \\
        Andrews \etal \cite{Andrews2017}            & Cargo          & Anomaly Detection                       & Train CNN features with Random Split Trees                              \\
        Jaccard \etal \cite{Jaccard2016b}           & Cargo          & Object Classification                   & oBIF+RF for non-empty cargo detection, followed by CNN for car detection                              \\
        Jaccard \etal \cite{Jaccard2016}            & Cargo          & Object Classification                   & CNN from scratch outperforms RF                               \\
        Rogers \etal \cite{Rogers2017a}             & Cargo          & Object Classification                   & Evaluation of high and low energy x-ray imagery                              \\
        Caldwell \etal \cite{Caldwell2017}          & Cargo, Baggage & Object Classification                   & Transferability between domains                              \\
        Yuan and Gui \cite{Yuan2018}                & Tera Hertz     & Object Classification                   & Two-stage. Classify from RGB, then Tera-Hertz images.                              \\
        Zhao \etal \cite{Zhao2018}                  & Baggage        & Image Generation,                       & Generate X-ray objects via GAN, and classify with CNN                            \\
                                                    &                & Object Classification                   &                                  \\
        Yang \etal \cite{Yang2019}                  & Baggage        & Image Generation                        & Generate X-ray objects via GAN, and classify with CNN                         \\
                                                    &                & Object Classification                   &                                  \\
        Miao \etal \cite{Miao2019}                  & Baggage        & Object Classification                   & with class-balanced hierarchical refinement                              \\
        Morris \etal \cite{Morris2018}              & Baggage        & Object Classification                   & Region-based detection with Z-effective                              \\
        Ak\c{c}ay and Breckon \cite{Akcay2017}      & Baggage        & Object Detection                        & Object Detection, Faster-RCNN is the best.                              \\
        Liang \etal \cite{Liang2018}                & Baggage        & Object Detection                        & RFCN is the best. Multi-view outperforms single view.                              \\
        Liang \etal \cite{Liang2019} & Baggage & Object Detection & Explores various detection algorithms, F-RCNN with Inception ResNet v2 achieves the highest performance \\
        Steitz \etal \cite{Steitz2018}              & Baggage        & Object Detection                        & F-RCNN with multi view pooling is superior to single view only.                              \\
        Liu \etal \cite{Liu2018}                    & Baggage        & Object Detection                        & YOLOv2 achieves real time performance.                              \\
        Xu \etal \cite{Xu2018}                      & Baggage        & Object Detection                        & Localizes the threat material from the X-ray images via attention mechanisms   \\
        Islam \etal \cite{Islam2018}                & Baggage        & Object Detection                        & track passengers and their belongings in airports while passing X-ray security checkpoints                              \\
        Liu \etal \cite{Jinyi2019}  & Baggage & Object Detection & Foreground object segmentation via material info, followed by a F-RCNN \\
        Gauss \etal \cite{Gaus2019a}                & Baggage &Object Detection & F-RCNN to investigate the tranferrability between various X-ray scanners. \\ 
        Cui and Oztan \cite{Cui2019a} & Baggage & Object Detection & RetinaNet trained on a TIP dataset achieves considerable faster detection than sliding window CNN. \\
        Hassan \etal \cite{Hassan2019} & Baggage & Object Detection & RoI are extracted via cascaded multiscale structure tensors, which are then classified via a CNN \\
        Bhowmik \etal \cite{Bhowmik2019} & Baggage & Object Detection & Explores the generalisation capability of the models trained on TIP datasets. \\
        Andrews \etal \cite{Andrews2016}            & Cargo          & Anomaly Detection                       & Fusion of the raw-input and residual error with feature encoding from the hidden layers.  \\
        Ak\c{c}ay \etal \cite{Akcay2018b}           & Baggage        & Anomaly Detection                       & encoder- decoder-encoder sub-networks. Minimize latent vector and image space.                              \\
        Ak\c{c}ay \etal \cite{Akcay2019}            & Baggage        & Anomaly Detection                       & Use of skip connections. Minimize latent vector in the discriminator network.                              \\
        Griffin \etal \cite{Griffin2019}            & Baggage        & Anomaly Detection                       & Feature Extraction with CNN, then train with Gaussian model.             \\ 
        Gauss \etal \cite{Gaus2019b} & Baggage & Object Segmentation & Mask-RCNN to segment RoI, and CNN classification for anomaly detection \\
        Bhowmik \etal \cite{Bhowmik2019a} & Baggage & Object Segmentation & Mask-RCNN to segment RoI, superpixel for sub-component level analysis, fine-grained CNN for classification \\
        An \etal \cite{An2019a} & Baggage & Object Segmentation & Dual attention mechanism within an encoder decoder segmentation network. \\ 
        \bottomrule
        \end{tabular}%
        }
        \caption{Overview of deep learning approaches applied within X-ray security imaging.}
        \label{tab:overview.of.dl.applications}
    \end{table}
    \end{landscape}
    
Another anomaly detection algorithm \cite{Griffin2019} (i) first extract the feature of the normal images from Inception v3 \cite{Szegedy2017} alike network, (ii) subsequently trains a multivariate Gaussian model to capture the normal distribution of \hyperref[datasets.ffob]{CAST} dataset. Anomaly score of a test sample is based on its likelihood that is relative to the model, which overall yields $92.5\%$ AUC.
\section{Discussion and Future Directions}
\label{sec:discussion}

Despite the promising performance of the proposed approaches, there are still some identifiable limitations. This section discusses the challenges and future directions based on the weaknesses and strengths of the current approaches presented in this paper and the broader literature including concurrent work to that presented here.

\paragraph{Dataset}
Although the use of transfer learning improves the performance of small X-ray datasets, the lack of large datasets limits contemporary deep model training. Relatively large datasets in the field such as \hyperref[datasets.sixray]{SIXray}, \hyperref[datasets.ffob]{FFOB} are highly biased towards certain classes, limiting to train reliable supervised methods. Hence, it is essential to build large, homogeneous, realistic and publicly available datasets, collected either by (i) manually scanning numerous bags with different objects and orientations in a lab environment or (ii) generating synthetic datasets via contemporary algorithms. 

There are advantages and disadvantages of both methods. Although manual data collection enables to gather realistic samples with the flexibility to produce any combination, it is rather expensive, requiring tremendous human effort and time.

Synthetic dataset generation, on that hand, is another method, currently achieved by TIP \cite{Rogers2016, Mery2017b} or GAN \cite{Zhao2018, Yang2019}. A recent study \cite{Bhowmik2019} empirically demonstrates that using a TIP dataset for a detection task adversely impacts the detection performance on real examples. In future work, therefore, more advanced algorithms such as image translation or domain adaptation \cite{Isola2016, Zhu2017} could be considered such that the model would learn to translate between benign and threat domains, which overall would yield superior projection/translation to TIP.

The literature has also seen another type of synthetic datasets generated by GAN algorithms. The limitation of current GAN datasets \cite{Zhao2018, Yang2019}, however, is that the models are currently incapable of producing full X-ray images. Moreover, the quality of the generated images is far from being realistic. Further studies, taking these issues into account, will need to be undertaken. It might be feasible to create more realistic X-ray images by using contemporary GAN algorithms \cite{Karras2019a}.

\paragraph{Exploiting Multiple-View Information}
Existing research recognizes the critical role played by multiple-view imagery, especially when the detection of an object from a particular viewpoint is challenging \cite{Michel2009, Steitz2018, Liang2018}.

Few studies \cite{Liang2018, Steitz2018, Isaac-Medina2020} investigate utilizing multiple-view integration inside/outside a CNN. Despite the incremental performance improvement reported, further work is required to investigate other possible ways to utilise multiple-view imagery better.


\paragraph{Domain Adaptation between X-ray Scanners}
As pointed out in \cite{Caldwell2017,Gaus2019a}, transferring models between different scanners could be challenging due to the unknown intrinsics of the scanners. Future work would utilize domain adaptation \cite{Zhu2017}, where the source domain contains images from one scanner, and the target domain would be of another X-ray scanner. Training with even unbalanced datasets would learn the intrinsic, and map from one to the other. 

\paragraph{Improving Unsupervised Anomaly Detection Approaches}
The performance of the current anomaly detection algorithms presented in Section \ref{subsec:dl.applications.anomaly.detection} is somewhat limited to be deployed for a real-world scenario. Therefore, more research on this topic needs to be undertaken to design better reconstruction techniques that thoroughly learn the characteristics of the normality from which the abnormality would be detected.

\paragraph{Use of the Material Information}
In dual-energy X-ray systems attenuation between high and low energies yields a unique value for different materials, which could be utilized further for more accurate object classification/detection \cite{Chen2007, Fu2010}. Even though recent research \cite{Morris2018,Rogers2017a} have examined the use of material information, the research outcome present inconsistent results. Hence, a further study thoroughly investigating the material information is suggested.

\section{Conclusion}
\label{sec:conclusion}
This paper taxonomises conventional machine and modern deep learning algorithms utilised within X-ray security imaging. Traditional approaches are sub-categorised based on computer vision tasks such as image enhancement, threat image projection, object segmentation, feature extraction, object classification, and detection. Review of the deep learning approaches includes classification, detection, segmentation and unsupervised anomaly detection algorithms applied within the field. 

Based on this review, several conclusion can be drawn for the future directions of the field. Despite the recently emerging datasets, the lack of large, well-balanced datasets limits the design of deep learning algorithms that are generalisable enough to be deployed in a real-time environment. Besides, since the public datasets are mostly from various machines with different intrinsic, the use of domain adaptation techniques could improve the generalisation capability of the algorithms.  

Unlike the abundance of studies in conventional machine learning, most of the recent approaches do not fully utilise X-ray imaging such as multiple-view geometry and high-low energy. Despite the existence of a few studies, there is room for further research. Moreover, further research in unsupervised learning could further utilise the existing X-ray datasets that are not labelled and not in use.

Overall, this paper reviews the strengths and weaknesses of the current techniques, and provides a thorough discussion for open challenges and envisions the future directions of the field.




{   
    \bibliographystyle{lib/model1-num-names}
    \singlespacing    
    \small
    \bibliography{ref/references.bib}
}







\clearpage
\end{document}